# Equivalent Constraints for Two-View Geometry: Pose Solution/Pure Rotation Identification and 3D Reconstruction

Qi Cai, Yuanxin Wu, Lilian Zhang and Peike Zhang

**Abstract**—Two-view relative pose estimation and structure reconstruction is a classical problem in computer vision. The typical methods usually employ the singular value decomposition of the essential matrix to get multiple solutions of the relative pose, from which the right solution is picked out by reconstructing the three-dimension (3D) feature points and imposing the constraint of positive depth. This paper revisits the two-view geometry problem and discovers that the two-view imaging geometry is equivalently governed by a Pair of new Pose-Only (PPO) constraints: the same-side constraint and the intersection constraint. From the perspective of solving equation, the complete pose solutions of the essential matrix are explicitly derived and we rigorously prove that the orientation part of the pose can still be recovered in the case of pure rotation. The PPO constraints are simplified and formulated in the form of inequalities to directly identify the right pose solution with no need of 3D reconstruction and the 3D reconstruction can be analytically achieved from the identified right pose. Furthermore, the intersection inequality also enables a robust criterion for pure rotation identification. Experiment results validate the correctness of analyses and the robustness of the derived pose solution/pure rotation identification and analytical 3D reconstruction.

---

## 1 INTRODUCTION

It is well known that the relative pose and 3D points can be generally recovered up to a scale from two views. The relationship of the image point pairs in two views is well described by two-view imaging geometry in Fig. 1. For the uncalibrated camera, it is specified as the fundamental matrix whose properties have been studied by Beardsley and Zisserman [1] and Vieville and Lingraud [2]. Hartley [3, 4] proposes algorithms for uncalibrated camera pose estimation. For the calibrated camera, the essential matrix was first introduced to the computer vision field by Longuet-Higgens [5], which provides a way to get the solution to the relative pose. Huang, Faugeras, Maybank and Hartley [6-9] have extensively studied the properties of the essential matrix.

The essential matrix is subject to the constraint $EE^T E - Tr(EE^T)E/2 = 0$ [6], which can be used in degenerate cases with less than eight point pairs. The degenerate cases are solved by Huang & Shim [10], Maybank [8], Nister [11] and Stewenius [12]. Usually the essential matrix linearly estimated using the normalized points would not be the optimum, and there exist nonlinear methods to refine the pose to fulfill the above constraint, as described in [13-15]. When a proper essential matrix is achieved, it can be used to get multiple solutions to the camera relative pose. The popular method is based on the singular value decomposition (SVD) of the essential matrix, which originated from the proof in Huang [7]. The SVD method was further developed and summarized in Hartley [9, 13] and Wang [16]. The solutions are used to reconstruct the 3D points by triangulation and it is only the right solution that yields positive depth [5, 17]. It has been believed that the SVD method requires non-zero translation [18, 19]. Recently, Kneip [20] puts forward a new nonlinear rotation constraint independent of translation and uses the Grobner basis method to solve the rotation for the 5-point case. That work also raises a new inequality to disambiguate the rotation matrix.

This paper is motivated by answering the following three questions:

1. How to explain the multiple pose solutions to the essential matrix equation from the aspect of solving equation?

2. How to explain the experiment phenomenon that the rotation part can still be recovered from the essential matrix equation in the pure rotation case?

3. Is it possible to directly identify the right pose solution without 3D reconstruction?

The main contribution of the paper is multiple-fold: 1) The two-view imaging geometry is found to be equivalently governed by a Pair of Pose-Only (PPO) constraints: the same-side constraint and the intersection constraint; 2) the basic two-view imaging equation is for the first time formulated as a function of pure pose, independent of 3D feature point coordinates. The relative depth information

- Qi Cai and Peike Zhang are with Shanghai Key Laboratory of Navigation and Location-based Services, School of Electronic Information and Electrical Engineering, Shanghai Jiao Tong University, Shanghai, China, 200240; and School of Aeronautics and Astronautics, Central South University, Changsha, Hunan, China, 410083. E-mail: {qicaiCN@gmail.com; zhangpeike@csu.edu.cn}.
- Yuanxin Wu is with Shanghai Key Laboratory of Navigation and Location-based Services, School of Electronic Information and Electrical Engineering, Shanghai Jiao Tong University, Shanghai, China, 200240. E-mail: yuanx_wu@hotmail.com.
- Lilian Zhang is with College of Mechatronics and Automation, National University of Defense Technology, Changsha, China, 410073. E-mail: lilianzhang@nudt.edu.cn.

is totally parameterized by relative pose; 3) The complete pose solutions to the essential matrix equation are explicitly derived, which explains that the orientation can still be recovered in the pure rotation case; 4) Aided by two inequalities derived from the same-side and intersection constraints, the right pose solution can be provably identified without resorting to 3D reconstruction. In other words, the pose identification is decoupled from 3D reconstruction, while 3D reconstruction can be analytically achieved from the identified right pose; 5) A robust criterion of pure rotation identification is derived from the intersection inequality.

The paper is organized as follows. Section 2 raises the PPO (same-side and intersection) constraints and proves their equivalency to the basic two-view imaging geometry. Section 3 derives the pose solutions to the essential equation. Section 4 simplifies the PPO constraints yielding the same-side inequality and the intersection inequality to help identify the right pose solution without 3D reconstruction of feature points. A pure rotation identification method is also derived from the intersection inequality. Section 5 reports the experiment results for pose solution/pure rotation identification and analytical reconstruction. Section 6 concludes the paper.

A number of frequently-used equalities of matrices, vectors and products are presented in Appendix for easy reference.

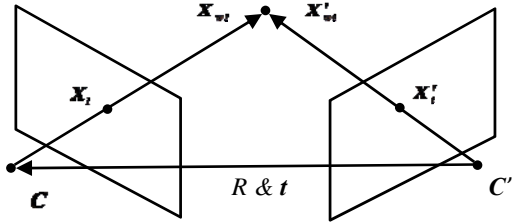

Fig 1. Sketch of two-view imaging geometry.

## 2 TWO-VIEW IMAGING GEOMETRY AND EQUIVALENT CONSTRAINTS

### 2.1 Two-View Geometry

The two-view imaging geometry is illustrated in Fig. 1. Suppose the point set $P = \{X_i = (x_i, y_i, 1)^T \mid i = 1, \ldots, m\}$ and the corresponding point set $P' = \{X'_i = (x'_i, y'_i, 1)^T \mid i = 1, \ldots, m\}$ are the sets of normalized image point pairs, while $P_w = \{X_{wi} = (x_{wi}, y_{wi}, z_{wi})^T \text{ or } X'_{wi} = (x'_{wi}, y'_{wi}, z'_{wi})^T \mid i = 1, \ldots, m\}$ are the projected 3D world feature points. Throughout the paper, we use the normalized image coordinates, unless explicitly stated otherwise.

The world point coordinates in the two camera frames, centered at $C$ and $C'$ respectively, are characterized by

$$X'_{wi} = RX_{wi} + t \qquad (1)$$

where $R$ and $t$ are respectively the true rotation matrix and translation vector between the two camera frames. It is readily apparent that the two normalized image point sets are related by

$$z'_{wi} X'_i = z_{wi} RX_i + t \qquad (2)$$

which can be rewritten as *the two-view imaging equation*

$$X'_i = \lambda_i (RX_i + s_i t) \qquad (3)$$

where the depth-related factors $\lambda_i \triangleq z_{wi}/z'_{wi} \in \Re^+$ and $s_i \triangleq 1/z_{wi} \in \Re^+$. The equation is important to almost all geometric computer vision problems.

Left multiplying (3) by $X'^T_i [t]_\times$ removes the depth-related factors [18]

$$0 = X'^T_i [t]_\times X'_i = \lambda_i X'^T_i [t]_\times RX_i \qquad (4)$$

which is right the well-known essential matrix equation [13]

$$X'^T_i E X_i = 0 \qquad (5)$$

where $E = [t]_\times R$ and $[t]_\times$ is the skew-symmetric matrix formed by $t$. Note that the development from (3) to (5) is not reversible, that is to say, some information has been lost as will be shown below. If the translation vector $t = 0$, the essential matrix $E$ reduces to zero or arguably is not well defined, but from the viewpoint of equation, (5) still exists with non-zero solutions, as discussed in next section.

Referring to Fig. 1, (5) depicts the *co-planar relationship* of the projection rays of the two views for the i-th point pair ($X'_i$ and $RX_i$) and the translation vector ($t$). *Note that it embeds the imaging geometry of two views, but loses the important imaging relationship among the translation vector and two projection rays.* For example, the two projection rays always point to the same side of the translation vector [20], as shown in Fig. 1. However, the obtained benefit of the coplanar relationship (5) is that the depth information ($\lambda_i$ and $s_i$) is isolated from pose and it is linear in $E$ which can be exploited to get pose solutions by linear methods like SVD.

Let multiplying (3) by $t^T [X'_i]_\times$ leads to

$$0 = \lambda_i t^T [X'_i]_\times (RX_i + s_i t) = \lambda_i t^T [X'_i]_\times RX_i$$
$$\Rightarrow t^T (X'_i \times RX_i) = 0 \qquad (6)$$

It means all vectors $\{X'_i \times RX_i \mid i = 1, \ldots, m\}$ lie on a plane with the normal vector $t$. Denote $B = (X'_1 \times RX_1, \ldots, X'_m \times RX_m)$, (6) is equivalent to $rank(B) = 2$. This is exactly the rotation constraint independent of translation proposed by [20].

### 2.2 Equivalent PPO Constraints

Next we will present the PPO constraints governing the physical formation of two overlapped views, namely, the same-side constraint and the intersection constraint (see Fig. 2). Define $\theta_1$, $\theta_2$ and $\theta_3$ to denote the angles among the translation vector and the projection rays, and $\theta_i \in [0, \pi]$, $i = 1, 2, 3$. Hereafter in this section, the translation vector is divided into the unit-direction part and the magnitude part by $t = e_t \|t\|$, where the unit vector $e_t$ is the direction vector of *t*. Note that for zero translation, $\|t\|=0$ and $e_t$ is an arbitrary unit vector.

**Same-side Constraint.** *The projection rays should point to the same side of the translation vector, that is to say, the vector products of $t \times RX_i$ and $t \times X_i'$ are in the same direction.*

The same-side constraint can be mathematically represented as

$$\frac{t \times RX_i}{\|t \times RX_i\|} = \frac{t \times X_i'}{\|t \times X_i'\|}$$
$$\Rightarrow \frac{e_t \times RX_i}{\|e_t \times RX_i\|} = \frac{e_t \times X_i'}{\|e_t \times X_i'\|} \quad (7)$$

The first equality is not well-defined for the zero-translation case. For zero translation, $\theta_3=0$ and $\theta_2=\theta_1$, and as a result (7) reduces to

$$\frac{e_t \times RX_i}{\|RX_i\|} = \frac{e_t \times X_i'}{\|X_i'\|} \Rightarrow e_t \times \left(\frac{RX_i}{\|RX_i\|} - \frac{X_i'}{\|X_i'\|}\right) = 0 \quad (8)$$

which means $RX_i/\|RX_i\| = X_i'/\|X_i'\|$, as $e_t$ is an arbitrary unit vector. In other words, the projection rays $X_i'$ and $RX_i$ are collinear in the same direction, so the second equality of (7) is a constraint for the same direction as well. That is to say, all translation cases are well defined by the second equality of (7).

**Intersection Constraint.** *The projection rays intersect with each other only if $\theta_2=\theta_1+\theta_3$ is satisfied.*

The three angles in Fig. 2 are obviously related to the vector inner products by

$$\cos\theta_1 = \frac{X_i'^T t}{\|X_i'\|\|t\|}, \cos\theta_2 = \frac{(RX_i)^T t}{\|X_i\|\|t\|}, \cos\theta_3 = \frac{X_i'^T RX_i}{\|X_i'\|\|X_i\|} \quad (9)$$

Note the arccosine function's value field is $[0,\pi]$. Taking cosine on both sides, the geometric constraint $\theta_2=\theta_1+\theta_3$ is equivalent to

$$\cos\theta_2 = \cos\theta_1 \cos\theta_3 - \sin\theta_1 \sin\theta_3 \text{ s.t. } \theta_1+\theta_3 \leq \pi \quad (10)$$

With (9), the equality of (10) can be rewritten as

$$\frac{(RX_i)^T t}{\|X_i\|\|t\|} = \frac{X_i'^T t}{\|X_i'\|\|t\|} \frac{X_i'^T RX_i}{\|X_i'\|\|X_i\|} - \frac{\|X_i' \times t\|}{\|X_i'\|\|t\|} \frac{\|X_i' \times RX_i\|}{\|X_i'\|\|X_i\|} \quad (11)$$

The inequality constraint of (10) is equivalent to $\cos\theta_1 \geq \cos(\pi - \theta_3)$, and we have

$$\frac{X_i'^T t}{\|X_i'\|\|t\|} + \frac{X_i'^T RX_i}{\|X_i'\|\|X_i\|} \geq 0 \quad (12)$$

Collectively, the intersection constraint is rewritten as

$$\begin{cases} \dfrac{(RX_i)^T t}{\|X_i\|\|t\|} = \dfrac{X_i'^T t}{\|X_i'\|\|t\|} \dfrac{X_i'^T RX_i}{\|X_i'\|\|X_i\|} - \dfrac{\|X_i' \times t\|}{\|X_i'\|\|t\|} \dfrac{\|X_i' \times RX_i\|}{\|X_i'\|\|X_i\|} \\ \dfrac{X_i'^T t}{\|X_i'\|\|t\|} + \dfrac{X_i'^T RX_i}{\|X_i'\|\|X_i\|} \geq 0 \end{cases} \quad (13)$$

or alternatively,

$$\begin{cases} \|X_i' \times e_t\|\|X_i' \times RX_i\| = X_i'^T RX_i X_i'^T e_t - \|X_i'\|^2 X_i^T R^T e_t \\ \|X_i\|X_i'^T e_t + X_i'^T RX_i \geq 0 \end{cases} \quad (14)$$

Similarly for the zero-translation case, (13) is not well-defined. In such a case, however, that the inequality in (14) is always true for any $e_t$ means $X_i^T R X_i \geq \max(-\|X_i\|X_i'^T e_t)$ $= \max(-\|X_i\|\|X_i'\|\cos\theta_1) = \|X_i\|\|X_i'\|$. As $X_i^T R X_i \leq \|X_i\|\|X_i'\|$, we can derive that $X_i^T R X_i = \|X_i\|\|X_i'\|$. This happens only when the projection rays $X_i'$ and $RX_i$ are collinear in the same direction, i.e., $\theta_3=0$. It shows that the inequality in (14) also gives a constraint for the same direction. We can easily verify that the equality in (14) becomes identity when motion is pure rotation. It is exactly what the intersection constraint means for the zero-translation case. That is to say, all translation cases are well defined by (14).

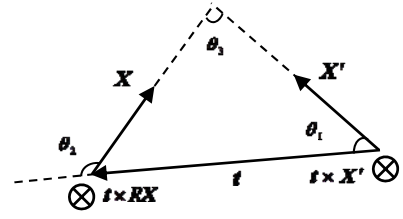

Fig 2. Two new geometric constraints: same-side constraint and intersection constraint.

**Proposition 1**: The two-view imaging equation is equivalent to the combination of the same-side and intersection constraints, namely, (7)+(14) $\Leftrightarrow$ (3).

**Proof of Sufficiency:**
The same-side constraint can be rewritten as

$$h \triangleq e_t \times wRX_i = e_t \times X_i' \quad (15)$$

where $w \triangleq \|e_t \times X_i'\|/\|e_t \times RX_i\|$. Define $q = h \times e_t$, $X_i'$ can be expressed as a linear combination of $t$ and $q$

$$X_i' = ae_t + bq \quad (16)$$

With (15),

$$\begin{aligned} q &= h \times e_t \\ &= (e_t \times wRX_i) \times e_t \\ &= wRX_i - w(X_i^T R^T e_t)e_t \end{aligned} \quad (17)$$

From the second line to the third line, the vector product equality for any triple vectors is used (see Appendix). Substituting (16) into (15) gives

$$h = e_t \times (ae_t + bq) = be_t \times q = bh \quad \Rightarrow b=1 \quad (18)$$

Considering (16)-(18), we have

$$\begin{aligned} X_i' &= ae_t + q \\ &= ae_t + \left(wRX_i - w(X_i^T R^T e_t)e_t\right) \\ &= wRX_i + (a - wX_i^T R^T e_t)e_t \end{aligned} \quad (19)$$

Contrasting (3) and (19), we will prove that $w=\lambda_i$ and $(a-wX_i^T R^T e_t) = \lambda_i s_i\|t\|$.

As for $w$,

$$w = \frac{\|e_t \times X_i'\|}{\|e_t \times RX_i\|} = \frac{z_{wi}}{z'_{wi}} \frac{\|e_t \times X_{wi}'\|}{\|e_t \times RX_{wi}\|} = \frac{z_{wi}}{z'_{wi}} = \lambda_i \quad (20)$$

where $\|e_t \times X_{wi}'\|$ and $\|e_t \times RX_{wi}\|$ both describe the distance from the 3D point to the translation baseline.

In order to prove $(a - wX_i^T R^T e_t) = \lambda_i s_i\|t\|$, we must take into account of the intersection constraint. Define the left side of the equality in (14) as $w_2 \triangleq \|e_t \times X_i'\| \|X_i' \times RX_i\|$. Substituting (16) and (18) into the equality in (14), we have

$$w_2 = ae_t^T RX_i q^T e_t + aq^T RX_i + q^T RX_i q^T e_t \\ -2ae_t^T q X_i^T R^T e_t - q^T q X_i^T R^T e_t \quad (21)$$

Considering (17), we have

$$q^T e_t = 0$$
$$q^T RX_i = wX_i^T X_i - w(X_i^T R^T e_t)^2 \quad (22)$$
$$q^T q = w^2 X_i^T X_i - w^2 (X_i^T R^T e_t)^2$$

So, (21) reduces to

$$w_2 = (a - wX_i^T R^T e_t) w(X_i^T X_i - (X_i^T R^T e_t)^2) \quad (23)$$

Define $w_3 \triangleq w(X_i^T X_i - (X_i^T R^T e_t)^2)$, then we have

$$(a - wX_i^T R^T e_t) = \frac{w_2}{w_3} \quad (24)$$

Substituting into (19), we get

$$X_i' = w\left(RX_i + \frac{w_2}{ww_3}e_t\right) \quad (25)$$

in which

$$\frac{w_2}{ww_3} = \frac{1}{w^2} \frac{\|e_t \times X_i'\| \|X_i' \times RX_i\|}{X_i^T X_i - (X_i^T R^T e_t)^2}$$
$$= \frac{1}{w^2} \frac{\|X_i'\|^2 \|X_i\| \sin\theta_1 \sin\theta_3}{\|X_i\|^2 - \|X_i\|^2 \cos^2\theta_2} \quad (26)$$
$$= \frac{1}{w^2} \frac{\|X_i'\|^2 \sin\theta_1 \sin\theta_3}{\|X_i\| \sin^2\theta_2}$$

Considering the relationship between normalized image points and the world coordinates of 3D points defined in (1) and (2), the above equation becomes

$$\frac{w_2}{ww_3} = \frac{1}{w^2} \frac{z_{wi}}{z'^2_{wi}} \frac{\|X'_{wi}\|^2 \sin\theta_1 \sin\theta_3}{\|X_{wi}\| \sin^2\theta_2}$$
$$= \frac{1}{w^2} \frac{z_{wi}}{z'^2_{wi}} \|t\| \frac{\|X'_{wi}\| \sin\theta_3}{\|t\| \sin\theta_2} \frac{\|e_t\| \|X'_{wi}\| \sin\theta_1}{\|e_t\| \|RX_{wi}\| \sin\theta_2} \quad (27)$$
$$= \frac{1}{w^2} \frac{z_{wi}}{z'^2_{wi}} \|t\| \frac{\|X'_{wi}\| \sin\theta_3}{\|t\| \sin\theta_2} \frac{\|e_t \times X'_{wi}\|}{\|e_t \times RX_{wi}\|}$$

As $\|X'_{wi}\| \sin\theta_3$ and $\|t\| \sin\theta_2$ both describe the distance from $C'$ to the projection ray $X_i$ and by using (20), it can be further reduced to

$$\frac{w_2}{ww_3} = \left(\frac{z'_{wi}}{z_{wi}}\right)^2 \frac{z_{wi}}{z'^2_{wi}} \|t\| = \|t\| s_i \quad (28)$$

Combing (20) and (26), (25) yields

$$X_i' = \lambda_i (RX_i + s_i \|t\| e_t) = \lambda_i (RX_i + s_i t) \quad (29)$$

**Proof of Necessity**:
Substitute (3) into the second equality of (7),

$$\frac{e_t \times X_i'}{\|e_t \times X_i'\|} = \frac{e_t \times \lambda_i (RX_i + s_i t)}{\|e_t \times \lambda_i (RX_i + s_i t)\|} = \frac{e_t \times RX_i}{\|e_t \times RX_i\|} \quad (30)$$

Similarly substitute (3) into the equality in (14),

$$X_i^{\prime T} RX_i X_i^{\prime T} e_t - \|X_i'\|^2 X_i^T R^T e_t \\ = \lambda_i (RX_i + s_i t)^T RX_i \lambda_i (RX_i + s_i t)^T e_t - \|\lambda_i (RX_i + s_i t)\|^2 X_i^T R^T e_t \quad (31)$$

Expanding the above equation yields

$$\lambda_i^2 (s_i X_i^T X_i t^T e_t - s_i X_i^T R^T t X_i^T R^T e_t) \\ = \lambda_i^2 s_i \|t\| (\|X_i\|^2 - (X_i^T R^T e_t)^2) \\ = \lambda_i^2 s_i \|t\| \|RX_i \times e_t\|^2 \quad (32) \\ = \|\lambda_i (RX_i + s_i t) \times e_t\| \|\lambda_i (RX_i + s_i e_t \|t\|) \times RX_i\| \\ = \|X_i' \times e_t\| \|X_i' \times RX_i\|$$

Substitute (3) into the inequality in (14),

$$\|X_i\| X_i^{\prime T} e_t + X_i^{\prime T} RX_i \\ = \lambda_i (RX_i + s_i t)^T (\|X_i\| e_t + RX_i) \quad (33) \\ = \lambda_i \left[\|X_i\| (\|X_i\| + e_t^T RX_i) + s_i \|t\| (\|X_i\| + e_t^T RX_i)\right] \geq 0$$

Q.E.D.

**Proposition 2**: The same-side constraint implies the co-planar relationship.
**Proof**.
Left multiply $\|t\| X_i^{\prime T}$ on both sides of the second equality of (7)

$$\|t\| X_i^{\prime T} \frac{e_t \times RX_i}{\|e_t \times RX_i\|} = \|t\| X_i^{\prime T} \frac{e_t \times X_i'}{\|e_t \times X_i'\|} \\ \Rightarrow X_i^{\prime T} EX_i = \|t\| \frac{\|e_t \times RX_i\|}{\|e_t \times X_i'\|} X_i^T (e_t \times X_i') = 0 \quad (34)$$

Q.E.D.

**Proposition 3**: The two-view imaging equation (3) can be rewritten as

$$X_i' = \frac{\|e_t \times X_i'\|}{\|e_t \times RX_i\|} RX_i + \frac{\|X_i' \times RX_i\|}{\|e_t \times RX_i\|} e_t \quad (35)$$

and the relative ratios of depth

$$\frac{z_{wi}}{\|t\|} = \frac{\|e_t \times X_i'\|}{\|X_i' \times RX_i\|} \text{ and } \frac{z'_{wi}}{\|t\|} = \frac{\|e_t \times RX_i\|}{\|X_i' \times RX_i\|} \quad (36)$$

**Proof**.
From (20),

$$\lambda_i = \frac{\|e_t \times X_i'\|}{\|e_t \times RX_i\|} \quad (37)$$

According to (20) and (26),

$$s_i = \frac{w_2}{\|t\| ww_3} = \frac{1}{\|t\| w^2} \frac{\|e_t \times X_i'\| \|X_i' \times RX_i\|}{X_i^T X_i - (X_i^T R^T e_t)^2} \\ = \frac{1}{\|t\|} \frac{\|e_t \times RX_i\|^2}{\|e_t \times X_i'\|^2} \frac{\|e_t \times X_i'\| \|X_i' \times RX_i\|}{\|e_t \times RX_i\|^2} \quad (38) \\ = \frac{1}{\|t\|} \frac{\|X_i' \times RX_i\|}{\|e_t \times X_i'\|}$$

Substitute (37) and (38) into (3),

$$X_i' = \frac{\|e_t \times X_i'\|}{\|e_t \times RX_i\|} \left(RX_i + \frac{1}{\|t\|} \frac{\|X_i' \times RX_i\|}{\|e_t \times X_i'\|} t\right) \\ \Rightarrow \|t\| \frac{\|e_t \times RX_i\|}{\|X_i' \times RX_i\|} X_i' = \|t\| \frac{\|e_t \times X_i'\|}{\|X_i' \times RX_i\|} RX_i + t \quad (39)$$

Comparing the above equation with (2) by coefficients, we have

$$z_{wi} = \|t\| \frac{\|e_t \times X'_i\|}{\|X'_i \times RX_i\|} \Rightarrow \frac{z_{wi}}{\|t\|} = \frac{\|e_t \times X'_i\|}{\|X'_i \times RX_i\|}$$

$$z'_{wi} = \|t\| \frac{\|e_t \times RX_i\|}{\|X'_i \times RX_i\|} \Rightarrow \frac{z'_{wi}}{\|t\|} = \frac{\|e_t \times RX_i\|}{\|X'_i \times RX_i\|}$$

(40)

Q.E.D.

**Proposition 4**: For all 3D points other than those on the baseline, $\|X'_i \times RX_i\| = 0$ if and only if $z_{wi}/\|t\| = +\infty$ and $z'_{wi}/\|t\| = +\infty$.

**Proof of Sufficiency**:
For the i-th 3D point,

$$\frac{z_{wi}}{\|t\|} = +\infty \text{ and } \frac{z'_{wi}}{\|t\|} = +\infty \qquad (41)$$

Considering (36), it results in $\|X'_i \times RX_i\| = 0$, because the numerators $\|e_t \times X'_i\|$ and $\|e_t \times RX_i\|$ are nonzeros for all 3D points other than those on the baseline.

**Proof of Necessity:**
If $\|X'_i \times RX_i\| = 0$, (36) implies (41).
Q.E.D.

**Proposition 5**: Pure rotation leads to $\|X'_i \times RX_i\| = 0$ for all point pairs, and an infinite 3D point or a point on the translation baseline only leads to $\|X'_i \times RX_i\| = 0$ for the corresponding point pair.

**Proof.**
From Proposition 4, for the case of pure rotation motion, $\|t\| = 0$ makes every point pair satisfy $\|X'_i \times RX_i\| = 0$. Similarly, for an infinite 3D point, $z_{wi}/\|t\| = +\infty$ and $z'_{wi}/\|t\| = +\infty$, the corresponding point pair satisfies $\|X'_i \times RX_i\| = 0$. And for a projected point on the baseline, $X'_i$ and $RX_i$ are collinear so that $\|X'_i \times RX_i\| = 0$.
Q.E.D.

Proposition 1 shows that the two-view imaging equation (3) is equivalent to the PPO constraints. In Section 4, we will see that the pair of constraints can be used to pick the right pose solution out of the multiple solutions obtained from (5). In Section 5, Proposition 3 will be used to analytically reconstruct the 3D point, and Proposition 4 will be used to discriminate infinite points or pure rotation motion.

## 3 COMPLETE POSE SOLUTIONS TO ESSENTIAL MATRIX EQUATION

The most well-known pose decomposition from the essential matrix is the SVD methods by Hartley[3] and others. But these methods are actually based on the implicit assumption of $t \neq 0$, namely, the essential matrix $E = [t]_\times R$ is well defined, so there exists an incorrect believing that in pure rotation case the essential matrix equation cannot be used to compute the rotation [18, 19].

This section is devoted to deriving all pose solutions to the essential equation (5) and fundamentally explaining the phenomenon that the orientation can be recovered from the case of pure rotation, which is incorrectly interpreted by [18] as the result of spurious translation caused by image coordinate noise. As a simplification of the two-view imaging equation, (5) might produce solutions that do not satisfy (3).

We first discuss the essential matrix solutions to the essential matrix equation (5) and then the pose solutions decomposed from the essential matrix solutions.

### 3.1 Essential Matrix Equation Solutions
In order to reduce the confusion, (5) is re-written as

$$X'^T_i Q X_i = 0 \Leftrightarrow (X^T_i \otimes X'^T_i) vec(Q) = 0 \qquad (42)$$

where $\otimes$ denotes the Kronecker product and $vec(\cdot)$ is the vectorization of a matrix (see Appendix). The popular method to solve $vec(Q)$ is the linear methods by restricting $vec(Q)$ to have unity length [13], in which the case of pure rotation has not been well considered.
Considering (3), we have

$$\left( X^T_i \otimes \begin{bmatrix} X^T_i & s_i \end{bmatrix} \right) \left( I_3 \otimes (R \quad t)^T \right) vec(Q) = 0 \qquad (43)$$

in which $\lambda_i$ is omitted as $\lambda_i > 0$. Then for the whole point set, we have

$$\begin{pmatrix} X^T_1 \otimes \begin{bmatrix} X^T_1 & s_1 \end{bmatrix} \\ \vdots \\ X^T_m \otimes \begin{bmatrix} X^T_m & s_m \end{bmatrix} \end{pmatrix} \left( I_3 \otimes (R \quad t)^T \right) vec(Q) = 0 \qquad (44)$$

Let

$$L \triangleq \begin{pmatrix} X^T_1 \otimes \begin{bmatrix} X^T_1 & s_1 \end{bmatrix} \\ \vdots \\ X^T_m \otimes \begin{bmatrix} X^T_m & s_m \end{bmatrix} \end{pmatrix}, \quad y \triangleq \left( I_3 \otimes (R \quad t)^T \right) vec(Q) \qquad (45)$$

Expanding the matrix $L$

$$L = \begin{pmatrix} x_1^2 & x_1 y_1 & x_1 & x_1 s_1 & y_1 x_1 & y_1^2 & y_1 & y_1 s_1 & x_1 & y_1 & 1 & s_1 \\ \vdots & \vdots & \vdots & \vdots & \vdots & \vdots & \vdots & \vdots & \vdots & \vdots & \vdots & \vdots \\ x_m^2 & x_m y_m & x_m & x_m s_m & y_m x_m & y_m^2 & y_m & y_m s_m & x_m & y_m & 1 & s_m \end{pmatrix} \qquad (46)$$

and (44) is equivalent to

$$Ly = 0 \qquad (47)$$

As columns 2, 3 and 7 are respectively equal to 5, 9 and 10, we have $rank(L) \leq 9$. Let $A = L(I_3 \otimes (R \quad t)^T)$, we have $rank(A) \leq 9$ and (44) can be written as $A vec(Q) = 0$. Therefore, when $m \geq 9$ the homogeneous equation (47) has three linearly independent special solutions

$$\xi_1 = (0 \quad 1 \quad 0 \quad 0 \quad -1 \quad 0 \quad 0 \quad 0 \quad 0 \quad 0 \quad 0 \quad 0)^T$$
$$\xi_2 = (0 \quad 0 \quad 1 \quad 0 \quad 0 \quad 0 \quad 0 \quad 0 \quad -1 \quad 0 \quad 0 \quad 0)^T \qquad (48)$$
$$\xi_3 = (0 \quad 0 \quad 0 \quad 0 \quad 0 \quad 0 \quad 1 \quad 0 \quad 0 \quad -1 \quad 0 \quad 0)^T$$

So the solution space of $y$ is given by

$$y = \left( I_3 \otimes (R \quad t)^T \right) vec(Q) = (a_1 \xi_1 + a_2 \xi_2 + a_3 \xi_3) \qquad (49)$$

where $a_1, a_2, a_3$ are real numbers. By using the Kronecker

product equality [21]
$$(I_3 \otimes (R \quad t)^T) vec(Q) = vec((R \quad t)^T Q) \quad (50)$$
(49) can be described in the form of matrix as
$$(R \quad t)^T Q = \begin{pmatrix} R^T Q \\ t^T Q \end{pmatrix} = \begin{pmatrix} 0 & -a_1 & -a_2 \\ a_1 & 0 & -a_3 \\ a_2 & a_3 & 0 \\ 0 & 0 & 0 \end{pmatrix} \quad (51)$$

Then we get two equations for $Q$
$$Q = R \begin{pmatrix} 0 & -a_1 & -a_2 \\ a_1 & 0 & -a_3 \\ a_2 & a_3 & 0 \end{pmatrix} = R[a]_\times \quad (52)$$
$$t^T Q = 0 \quad (53)$$

where $a \triangleq (a_3, -a_2, a_1)^T$ and $a$ can be determined without translation information from (52). Equation (52) indicates that the solutions to (42) are definitely essential matrices. For $t = 0$, it is obvious that $Q$ is still related to $R$, but its connection to $t$ would be gone. It can be derived from (52) that $(I_3 \otimes R^T) vec(Q) = vec([a]_\times)$ which means
$$\|a\| = \sqrt{a_1^2 + a_2^2 + a_3^2} = \frac{\sqrt{2}}{2} \|vec(Q)\| \quad (54)$$
as the matrix $I_3 \otimes R^T$ is orthogonal.

TABLE I. VALIDITY OF ROTATION MATRIX

| | | |
|---|---|---|
| $det(U_Q V_Q) = 1$ | $R = U_Q W_1^T V_Q^T \to \det(R) = 1$ | valid |
| | $R = U_Q W_2^T V_Q^T \to \det(R) = 1$ | valid |
| | $R = U_Q W_3^T V_Q^T \to \det(R) = -1$ | invalid |
| | $R = U_Q W_4^T V_Q^T \to \det(R) = -1$ | invalid |
| $det(U_Q V_Q) = -1$ | $R = U_Q W_1^T V_Q^T \to \det(R) = -1$ | invalid |
| | $R = U_Q W_2^T V_Q^T \to \det(R) = -1$ | invalid |
| | $R = U_Q W_3^T V_Q^T \to \det(R) = 1$ | valid |
| | $R = U_Q W_4^T V_Q^T \to \det(R) = 1$ | valid |

### 3.2 Complete Pose Solutions

According to (52), $Q$ and $[a]_\times$ have the same singular values. By SVD, the skew-symmetric matrix can be decomposed as $[a]_\times = U_\times \Sigma V_\times^T$, where $U_\times$, $V_\times^T$ are orthogonal matrices and $\Sigma = diag(\|a\|, \|a\|, 0)$ [21]. Then $Q$ can be written as
$$Q = R[a]_\times = R U_\times \Sigma V_\times^T \triangleq U_Q \Sigma V_Q^T \quad (55)$$
where $U_Q = R U_\times$ and $V_Q = V_\times$. Note that the SVD decomposition is not unique. Actually, to get $R$ and $t$, there is no need to get the exact form of $V_\times$. As $[a]_\times$ is a skew-symmetric matrix, one of the singular values is equal to zero and the corresponding singular value vector is $a$ [21]. Let $U_\times = (u_1, u_2, u_3)$ and $V_\times = (v_1, v_2, v_3)$, where $v_3 = \pm a/\|a\|$ and $v_1, v_2$ are the singular value vectors for the singular value $\|a\|$. It can be derived from $[a]_\times = U_\times \Sigma V_\times^T$ that

$$(u_1, u_2) = [a]_\times (v_1, v_2) \begin{pmatrix} \|a\| & 0 \\ 0 & \|a\| \end{pmatrix}^{-1}$$
$$= \frac{1}{\|a\|} ([a]_\times v_1, [a]_\times v_2) \quad (56)$$
$$= \left( \frac{a}{\|a\|} \times v_1, \frac{a}{\|a\|} \times v_2 \right)$$

The explicit relationship between $U_\times$ and $V_\times$ is complicated, due to different combinations of the determinant of $U_\times$ and $V_\times$. We next examine it case by case.

If $det(V_\times) = 1$,
  Case (1). $v_3 = a/\|a\|$ & $u_3 = v_3 \to U_\times = V_\times W_1$
  Case (2). $v_3 = -a/\|a\|$ & $u_3 = v_3 \to U_\times = V_\times W_2$
  Case (3). $v_3 = a/\|a\|$ & $u_3 = -v_3 \to U_\times = V_\times W_3$
  Case (4). $v_3 = -a/\|a\|$ & $u_3 = -v_3 \to U_\times = V_\times W_4$

If $det(V_\times) = -1$,
  Case (1). $v_3 = a/\|a\|$ & $u_3 = v_3 \to U_\times = V_\times W_2$
  Case (2). $v_3 = -a/\|a\|$ & $u_3 = v_3 \to U_\times = V_\times W_1$
  Case (3). $v_3 = a/\|a\|$ & $u_3 = -v_3 \to U_\times = V_\times W_4$
  Case (4). $v_3 = -a/\|a\|$ & $u_3 = -v_3 \to U_\times = V_\times W_3$

where $W_1 = \begin{pmatrix} 0 & -1 & 0 \\ 1 & 0 & 0 \\ 0 & 0 & 1 \end{pmatrix}$, $W_2 = \begin{pmatrix} 0 & 1 & 0 \\ -1 & 0 & 0 \\ 0 & 0 & 1 \end{pmatrix}$, $W_3 = \begin{pmatrix} 0 & -1 & 0 \\ 1 & 0 & 0 \\ 0 & 0 & -1 \end{pmatrix}$

and $W_4 = \begin{pmatrix} 0 & 1 & 0 \\ -1 & 0 & 0 \\ 0 & 0 & -1 \end{pmatrix}$. According to (55), the right rotation

is related to the non-unique SVD decomposition by
$$R = U_Q U_\times^T = U_Q W_i^T V_Q^T, \quad i = 1,2,3,4 \quad (57)$$
among which there exist invalid cases that can be easily removed by using the fact that $R$ is required to a rotation matrix, i.e., $det(R) = 1$, as shown in Table I. Additionally, let $U_Q = (u_{q1}, u_{q2}, u_{q3})$, it is obvious
$$(u_{q1}, u_{q2}, u_{q3}) = (R u_1, R u_2, R u_3) \quad (58)$$
Considering Table I and the above discussion of different cases, if $det(U_Q V_Q) = 1$
$$u_{q3} = R u_3 = \pm R v_3 = \pm R \frac{a}{\|a\|} \quad (59)$$

According to (53), the relationship between $a$ and $t$ can be known as
$$t^T Q = 0 \Rightarrow (R[a]_\times)^T t = (R^T t) \times a = 0 \quad (60)$$
So if $t \neq 0$, we have
$$e_t = \pm R a/\|a\| \quad (61)$$
Taking (59) into consideration, the translation vector can be derived as
$$\frac{t}{\|t\|} = e_t = \pm u_{q3} \quad (62)$$
In summary, the number of complete pose solutions to the essential matrix equation (42) is eight, which are now concluded as $[R|t]$, where $\{R = R_i | R_i \triangleq U_Q W_i^T V_Q^T, i=1,2,3,4\}$ and $\{t = t_1 \text{ or } t_2 | t_1 \triangleq u_{q3}, t_2 \triangleq -u_{q3}\}$.

They are consistent with those in the previous work of Wang et al. [16]. As $U_Q$ and $V_Q$ are usually obtained by the general SVD, $det(U_Q V_Q)$ may be 1 or -1. As shown in Table I, if $det(U_Q V_Q)=1$, $R_1$ and $R_2$ are valid; otherwise, $R_3$ and $R_4$ are valid. So in practical code implementation, we can reduce the number of potential pose solutions from eight to four by simply restricting $det(U_Q V_Q)=1$, namely, $[R_1|t_1]$, $[R_2|t_1]$, $[R_1|t_2]$ and $[R_2|t_2]$. Hartley et al. [13] proposed these four solutions by imposing the constraint $det(U_Q) = det(V_Q) = 1$ that falls in the scope of $det(U_Q V_Q)=1$. Therefore, the essential matrix equation (5) has no less than four pose solutions.

**Proposition 6**: For the case $t \neq 0$, $Q = \pm E\|a\|/\|t\|$ (or $Q$ is equivalent to $E$ up to scale); when $t = 0$, $E = 0$ and $Q$ has nothing to do with $E$.

**Proof**.
From (52) and (61), if $t \neq 0$

$$Q = R[a]_\times = \pm R\left[\frac{\|a\|}{\|t\|}R^T t\right]_\times = \pm \frac{\|a\|}{\|t\|}RR^T[t]_\times R = \pm \frac{\|a\|}{\|t\|}E \quad (63)$$

Q.E.D.

**Proposition 7**: For the case $t = 0$, the right rotation matrix $R$ can be obtained as one of the solutions; the translation estimate is always wrong.

**Proof**.
For the case of $t = 0$, the right rotation matrix can be obtained from (57) because (52) is still valid. However, the translation vector estimate satisfies $\hat{t} = \pm u_{q3} = \pm Ra/\|a\|$ according to (59). It has nothing to do with the true translation, namely, zero translation. As we know, the norm of $\hat{t}$ is always equal to 1.

Q.E.D.

## 4 IDENTIFY RIGHT SOLUTION WITHOUT 3D RECONSTRUCTION

This section will simplify the PPO equality constraints into linear inequality constraints and then use them to identify the right solution from the multiple solutions to the essential equation. Section 3 explores the theoretical pose solutions to the essential matrix equation (5). In applications, an estimate of the essential matrix $\hat{Q}$ is obtained as the solution of (42). Hereafter we use the hatted symbols to distinguish the estimate from the theoretical solution above. Then by the popular SVD method, four pose solutions can be obtained by imposing the constraint $det(U_{\hat{Q}} V_{\hat{Q}})=1$, namely, $[\hat{R}_1|\hat{t}_1]$, $[\hat{R}_2|\hat{t}_1]$, $[\hat{R}_1|\hat{t}_2]$ and $[\hat{R}_2|\hat{t}_2]$, where $\{\hat{R}_i = U_{\hat{Q}} W_i^T V_{\hat{Q}}^T | i=1,2\}$, $\hat{t}_1 = u_{q3}$ and $\hat{t}_2 = -u_{q3}$, as in [13, 16].

### 4.1 Same-Side Inequality Constraint

From the same-side constraint (7),

$$\left(\frac{e_t \times RX_i}{\|e_t \times RX_i\|}\right)^T \frac{e_t \times X_i'}{\|e_t \times X_i'\|} = 1 \quad (64)$$

It can be reduced to

$$(e_t \times RX_i)^T(e_t \times X_i') = \|e_t \times RX_i\|\|e_t \times X_i'\|$$
$$= X_i^T R^T [e_t]_\times^T [e_t]_\times X_i' > 0 \quad (65)$$

This is in spirit similar to the inequality to disambiguate the rotation matrix by Kneip [20], which is nonlinear in the rotation matrix. Note that this inequality has a quite wide threshold, namely $\|e_t \times RX_i\|\|e_t \times X_i'\|$, which is very helpful in robustly identifying the right solution, as shown in next section.

Considering (52) and (61), $[e_t]_\times^T [e_t]_\times = QQ^T/\|a\|^2$, so the constraint for a single point is formulated as

$$m_1(R) \triangleq X_i^{\prime T} Q Q^T R X_i > 0 \quad (66)$$

where $m_1(R)$ is a function of $R$ when others are known.

It can be checked that the above inequality is able to discriminate the right orientation. Given $\hat{Q}$ and $\hat{R} = \hat{R}_1 \text{ or } \hat{R}_2$,

$$X_i^{\prime T} \hat{Q}\hat{Q}^T \hat{R} X_i = X_i^{\prime T} U_{\hat{Q}} \Sigma\Sigma^T W_1^T V_{\hat{Q}}^T X_i \text{ or}$$
$$X_i^{\prime T} U_{\hat{Q}} \Sigma\Sigma^T W_2^T V_{\hat{Q}}^T X_i \quad (67)$$

As $\Sigma\Sigma^T W_1^T = -\Sigma\Sigma^T W_2^T$, we have

$$m_1(\hat{R}_1) = -m_1(\hat{R}_2) \quad (68)$$

i.e.

$$m_1(\hat{R}_1) m_1(\hat{R}_2) < 0 \quad (69)$$

The equation shows that the two orientation solutions yield $m_1(\hat{R})$ with opposite sign, so it is feasible to get the right orientation by using $m_1(\hat{R}) > 0$.

For all point pairs, the same-side linear inequality constraint can be collectively written as

$$M_1(\hat{R}) \triangleq \begin{pmatrix} X_1^T \otimes X_1^{\prime T} \\ \vdots \\ X_m^T \otimes X_m^{\prime T} \end{pmatrix} vec(\hat{Q}\hat{Q}^T \hat{R}) > 0 \quad (70)$$

which can be used to select the right rotation matrix.

### 4.2 Intersection Inequality Constraint

Since $\theta_2 = \theta_1 + \theta_3$ and $\theta_i \in [0, \pi]$, we have $\pi \geq \theta_2 > \theta_1 \geq 0$. It means

$$\cos^{-1}\frac{(RX_i)^T t}{\|X_i\|\|t\|} > \cos^{-1}\frac{X_i^{\prime T} t}{\|X_i'\|\|t\|} \quad (71)$$

i.e.

$$\frac{X_i^{\prime T} e_t}{\|X_i'\|} > \frac{X_i^T R^T e_t}{\|X_i\|} \quad (72)$$

It can be written as

$$m_2(R, t) \triangleq \left(\|X_i'\| X_i^T \quad \|X_i\| X_i^{\prime T}\right) \begin{pmatrix} -R^T \\ I_3 \end{pmatrix} e_t > 0 \quad (73)$$

For the translation solutions in Section 3, we can see that $m_2(R,t)$ is effective as $m_2(\hat{R},\hat{t}_1)m_2(\hat{R},\hat{t}_2)<0$. Actually, the sign of $m_2(R,t)$ is independent of the right or wrong R. Let $\hat{t}_1 = u_{q3}$, for $\hat{R}_1$ and $\hat{R}_2$, we have

$$\begin{pmatrix} -\hat{R}_1^T \\ I_3 \end{pmatrix}\hat{t}_1 = \begin{pmatrix} -(U_Q W_1^T V_Q^T)^T \\ I_3 \end{pmatrix} u_{q3} = \begin{pmatrix} -V_Q W_1 U_Q^T u_{q3} \\ u_{q3} \end{pmatrix}$$

$$= \begin{pmatrix} -V_Q W_1 \begin{pmatrix} 0 \\ 0 \\ 1 \end{pmatrix} \\ u_{q3} \end{pmatrix} = \begin{pmatrix} -v_{q3} \\ u_{q3} \end{pmatrix} \quad (74)$$

and

$$\begin{pmatrix} -\hat{R}_2^T \\ I_3 \end{pmatrix}\hat{t}_1 = \begin{pmatrix} -V_Q W_2 \begin{pmatrix} 0 \\ 0 \\ 1 \end{pmatrix} \\ u_{q3} \end{pmatrix} = \begin{pmatrix} -v_{q3} \\ u_{q3} \end{pmatrix} \quad (75)$$

So $m_2(\hat{R}_1,\hat{t}_1) = m_2(\hat{R}_2,\hat{t}_1)$ when $\hat{t}_1 = u_{q3}$ (or $\hat{t}_2 = -u_{q3}$). In conclusion, the intersection linear inequality constraint $m_2(R,t)$ can be described as $m_2(t)>0$.

For all point pairs, the intersection inequality constraint can be collectively written as

$$M_2(\hat{t}) \triangleq \begin{pmatrix} \|X_1'\|X_1^T & \|X_1\|X_1'^T \\ \vdots & \vdots \\ \|X_m'\|X_m^T & \|X_m\|X_m'^T \end{pmatrix} \begin{pmatrix} -\hat{R}^T \\ I_3 \end{pmatrix}\hat{t} > 0 \quad (76)$$

which can used to select the right translation vector.

In contrast to the robustness in identifying the rotation matrix by $M_1(\hat{R})>0$, finding the right translation vector by $M_2(\hat{t})>0$ in the case of large depth/translation ratio should be cautious. In these cases, according to Proposition 4, $\|X_i' \times RX_i\| \approx 0$ and $X_i' \approx RX_i\|e_t \times X_i'\|/\|e_t \times RX_i\| = \lambda_i RX_i$ from (35). $m_2(t)$ might be very near to zero because

$$m_2(t) = (-\|X_i'\|X_i^T R^T + \|X_i\|X_i'^T)e_t$$
$$\approx (-\|\lambda_i RX_i\|X_i^T R^T + \|X_i\|(\lambda_i RX_i)^T)e_t \quad (77)$$
$$= 0$$

This might cause some problem in translation identification in the case of nearly pure rotation. However, translation estimation under this case is of little significance. In this aspect, it would be very helpful for judging the significance of the obtained translation vector if the pure rotation case could be successfully identified.

### 4.3 Robust Relative Pose Algorithm Without 3D Reconstruction

Table II summarizes the algorithm of linear solutions of the essential matrix and how to robustly identify the right solution using the proposed two inequality constraints. Due to the effect of point matching error, there might exist some point pairs that violate the constraints. To improve the robustness against such errors, we could choose the solution as the right one with the maximum number of point pairs that satisfy $M_1(\hat{R})>0$ and $M_2(\hat{t})>0$.

TABLE II. POSE ESTIMATION WITHOUT 3D RECONSTRUCTION

| | |
|---|---|
| Step 1. | Given the linear solution matrix $Q$; |
| Step 2. | By SVD, $Q = U_Q \Sigma V_Q^T$. Choose $W_1$ or $W_2$ if $det(U_Q V_Q) = 1$. Otherwise, choose $W_3$ or $W_4$; |
| Step 3. | Use $\hat{t} = \pm u_{q3}$ and $\hat{R} = U_Q W_1^T V_Q^T$ or $U_Q W_2^T V_Q^T$ (or other methods) to get potential four pose solutions; |
| Step 4. | Count the number of positive entities in $M_1(\hat{R})$, denoted as $S_1(\hat{R})$, then the right rotation should be $\hat{R} = \max_{\hat{R}}\{S_1(\hat{R})\}$. Similarly, count the number of positive entities in $M_2(\hat{t})$, denoted as $S_2(\hat{t})$, then the right translation should be $\hat{t} = \max_{\hat{t}}\{S_2(\hat{t})\}$. |

### 4.4 Relative Rotation Identification

Propositions 4-5 prove the property of pure rotation and its difference from infinite points and baseline points (i.e., points on the translation baseline). In this section, R denotes the right rotation matrix identified by $M_1(\hat{R})>0$. To identify pure rotation, we might manually set a threshold on $\|X_i' \times RX_i\|$ for all point pairs. In other words, we can identify the pure rotation motion by

$$m_3(R) \triangleq \frac{\|X_i' \times RX_i\|}{\|X_i'\|\|X_i\|}, M_3(R) \triangleq avg\begin{pmatrix} \frac{\|X_1' \times RX_1\|}{\|X_1'\|\|X_1\|} \\ \vdots \\ \frac{\|X_m' \times RX_m\|}{\|X_m'\|\|X_m\|} \end{pmatrix} < \delta_\theta \quad (78)$$

where $\delta_\theta$ is a threshold and $avg(\cdot)$ is the average function. We identify the camera motion as pure rotation if $M_3(R)$ is greater than some-prescribed confidence level, e.g., 0.95. There exists a problem that $\delta_\theta$ needs to be manually tuned under different cases.

**Proposition 8**: For all 3D points other than those on the baseline, $\|X_i' \times RX_i\|=0$ if and only if $m_2(t)=0$.

**Proof of Necessity**:

If $\|X_i' \times RX_i\|=0$, according to Proposition 4 and referring to the development in (77), $X_i' = RX_i\|e_t \times X_i'\|/\|e_t \times RX_i\| = \lambda_i RX_i$ from (35). Then we have $m_2(t)=0$.

**Proof of Sufficiency**:

If $m_2(t)=0$, referring to Fig. 2, we have

$$m_2(t) = -\|X_i'\|X_i^T R^T e_t + \|X_i\|X_i'^T e_t = 0$$
$$\Rightarrow \|X_i\|\|X_i'\|\cos\theta_1 - \|X_i\|\|X_i'\|\cos\theta_2 = 0 \quad (79)$$
$$\Rightarrow \cos\theta_1 = \cos\theta_2$$

which means $\theta_1 = \theta_2$ as $\theta_1, \theta_2 \in [0,\pi]$ and the cosine function is monotonic. From the intersection constraint, the angle $\theta_3$ must be zero, namely, $\|X_i' \times RX_i\|=0$.

Q.E.D.

Note that the normal scenes will not contain infinite points and translation baseline points only. According to Propositions 4-5, Proposition 8 actually tells that $m_2(t)=0$ could be alternatively used as a criterion for identifying pure rotation.

## 5 TEST RESULTS

In our simulation tests, the 3D feature points are centered at known coordinates and subjected to uniform distributions, which are mapped to two views to generate image pair coordinates in pixels. For example, we use $U((0,0,d),(30,30,d))$ to denote the uniform distribution in a cube that is centered at $[0\ 0\ d]^T$ and of length 30, 30 and $d$ (in meters) along X, Y and Z axes, respectively. The extrinsic parameter matrix of the left view is constantly set to $(I_3 | 0_{3\times 1})$ and the intrinsic parameter matrix of two cameras is $\begin{pmatrix} 800 & 0 & 512 \\ 0 & 800 & 512 \\ 0 & 0 & 1 \end{pmatrix}$.

We generate $N_{pts}$ 3D points in $U((0,0,d),(30,30,d))$. To get $N_s$ poses of the right view, we generate the rotation matrices of which the corresponding Euler angles obey $N([0\ 0\ 0]^T, \text{diag}[5^2\ 5^2\ 20^2])$ in degrees, and the unit direction vector of translation in the OXY plane. Each translation length is given by $\|t\|=\alpha\|t\|_{\max}$, where the parallax factor $\alpha \in [10^{-3},1]$ and $\|t\|_{\max}=4.2$ denotes the maximum length of translation. Figure 3 demonstrates 3D points and the poses of right views for $d=30$, $N_{pts}=50$ and $N_s=10$.

Image points are obtained by projecting 3D points onto the pair of views and the Gaussian noise is added to the generated image pair coordinates, with the standard deviation varying between 0.1~5 pixels for $N_{std}$ times. $N_{mc}$ Monte Carlo runs are carried out for each case. In the following tests, we set ($N_s$, $N_{pts}$, $N_{std}$, $N_{mc}$)=(200, 50, 200, 30)

TABLE III. POSE SOLUTIONS FOR A PURE ROTATION CASE

| True Pose (Rotation in Euler angles, degree; Translation) | Four Pose Solutions |
|---|---|
| $\begin{pmatrix} 10 & -5 & 2 \\ 0 & 0 & 0 \end{pmatrix}$ | $\begin{pmatrix} 10 & -5 & 2 \\ 0.973 & 0.154 & -0.174 \end{pmatrix}$ |
| | $\begin{pmatrix} 10 & -5 & 2 \\ -0.973 & -0.154 & 0.174 \end{pmatrix}$ |
| | $\begin{pmatrix} -167.358 & 24.864 & 16.724 \\ 0.973 & 0.154 & -0.174 \end{pmatrix}$ |
| | $\begin{pmatrix} -167.358 & 24.864 & 16.724 \\ 0.973 & -0.154 & -0.174 \end{pmatrix}$ |

and $d=30$ m unless explicitly stated otherwise.

The experimental phenomenon that the rotation part can still be recovered from the essential matrix equation in the pure rotation case has been reported in very few literature, e.g., [18]. Table III lists the pose decomposition result in a case of pure rotation. As predicted in Section 4, we see that the rotation matrix can be still recovered, but the translation vector is wrong.

### 5.1 Pose Solution Identification Result

We define the discrepancy $\varepsilon_{i,j}^R$ between the identified rotation matrix and the true rotation matrix as

$$\varepsilon_{i,j}^R = \frac{1}{3} \left\| dcm2angle\left(R_i^T \hat{R}_{i,j}\right) \right\|_{L1} \tag{80}$$

where $R_i$ is the true rotation of the $i$-th two-view and $\hat{R}_{i,j}$ is the identified rotation matrix in the $j$-th Monte Carlo run. $dcm2angle(\cdot)$ means the Euler angle vector derived from the rotation matrix, and $\|\cdot\|_{L1}$ is the 1-norm of a vector. Similarly, define the discrepancy $\varepsilon_{i,j}^t$ between the identified translation vector and the true translation by

$$\varepsilon_{i,j}^t = \arccos\left( \frac{t_i^T \hat{t}_{i,j}}{\|\hat{t}_{i,j}\| \|t_i\|} \right) \tag{81}$$

Figure 4 plots the rotation and translation discrepancies (RMSE), averaged across Monte Carlo runs, as the function of noise standard deviation and the parallax factor $\alpha$. The result in Fig. 4(a) accords with Proposition 7 that we can still get good rotation estimates even when $\alpha$ approaches zero. Actually, the rotation identification is always right. However, as shown in Fig. 4(b), the estimated translation vector is susceptible to the noise standard deviation or the parallax factor. The translation discrepancy even reaches 90 degree at the right-bottom triangular area where the noise standard deviation is large or the parallax factor is small. This observation shows a fact that the translation estimate obtained from the essential equation is unsatisfactory in accuracy.

In order to investigate the robustness of the pose identification method proposed in Section 4, we compare it with that of the traditional method by

$$\varepsilon_{diff}^t = \varepsilon_{trad}^t - \varepsilon_{new}^t \tag{82}$$

where $\varepsilon_{trad}^t$ and $\varepsilon_{new}^t$ are translation discrepancies for the traditional method [13] and the new method, respectively, and $\varepsilon_{diff}^t$ is the difference of the two translation discrepancies, as shown in Fig. 5. In order to get a more intuitive result of relative advantage, we arguably average the results of all parallax factors for a definite noise standard deviation and vice visa, as shown in Figs. 6(a)-(b) for $d$=30m, 50m and 80m. If we regard Fig. 5 as a matrix, they respectively correspond to column averaging and row averaging. In Fig. 6, the translation discrepancy differences are almost all above zero, which shows that the new identification method performs more robustly. As shown in Fig. 6(a), the larger noise standard deviation becomes, the better robustness is. In Fig. 6(b), the trend is similar for small parallax

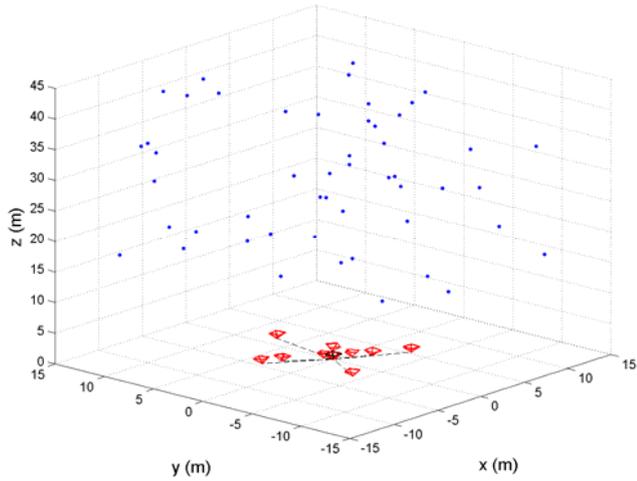

Fig 3. An example of synthetic 3D points and right views.

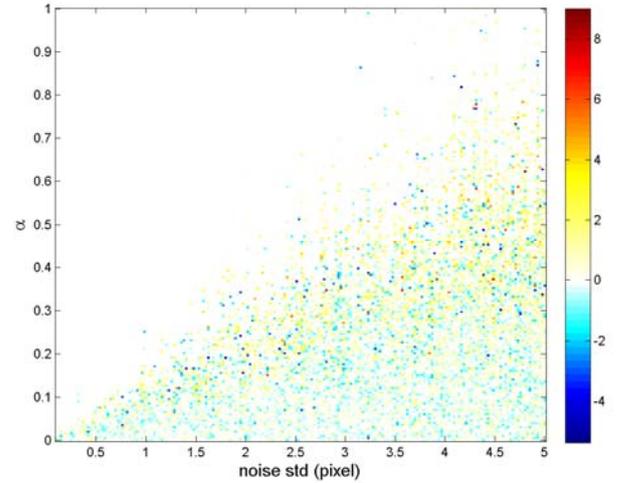

Fig 5. Translation discrepancy difference relative to traditional method (degree), as a function of parallax factor and noise standard deviation.

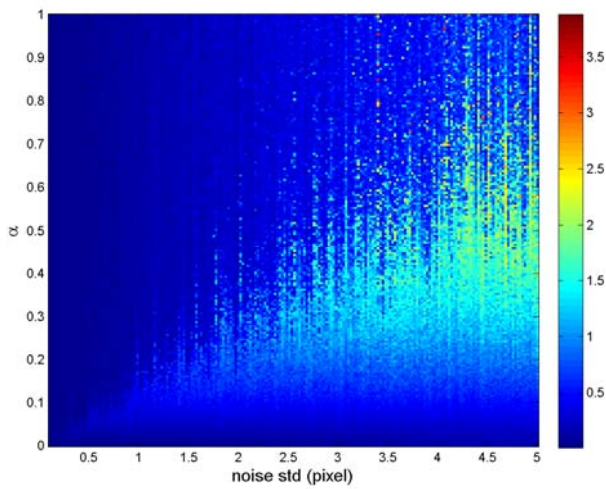

Fig 4(a). Rotation discrepancy (degree), as a function of parallax factor and noise standard deviation.

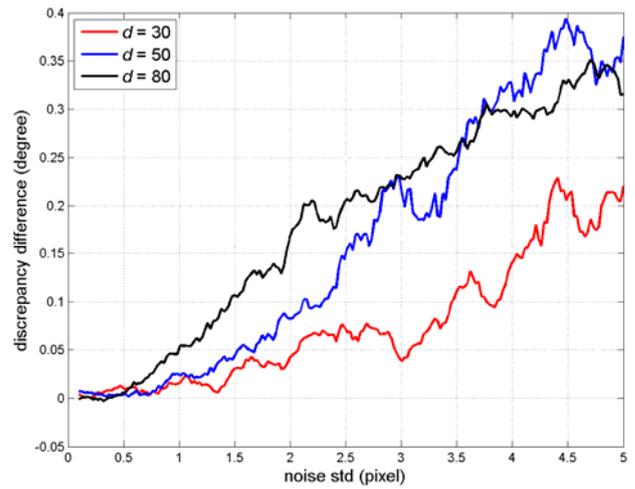

Fig 6(a). Translation discrepancy difference averaged across all parallax factors.

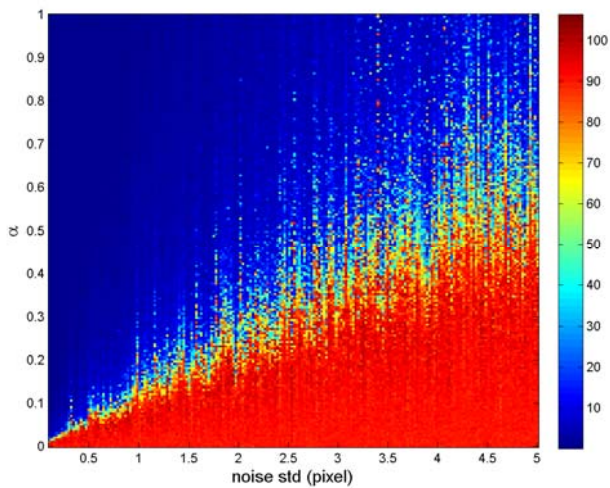

Fig 4(b). Translation discrepancy (degree), as a function of parallax factor and noise standard deviation.

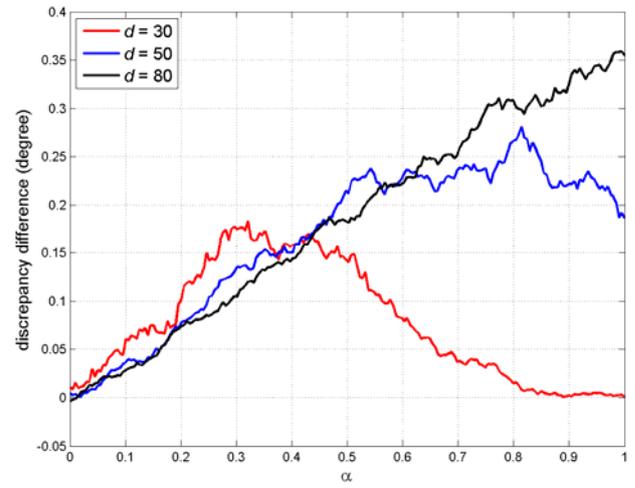

Fig 6(b). Translation discrepancy difference averaged across all noise standard deviations.

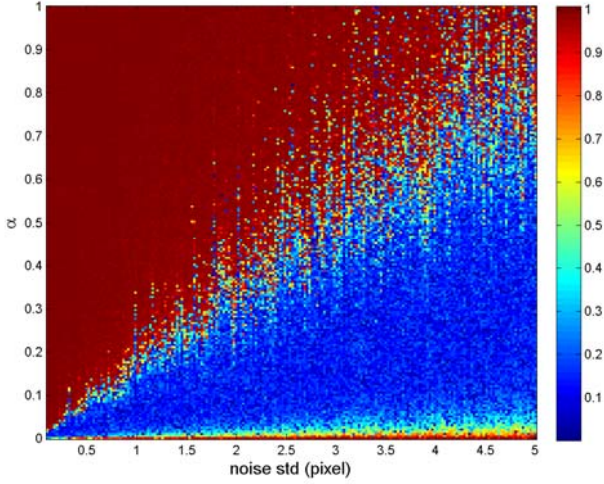

Fig 7. Reconstruction error ratio of analytical method relative to traditional method, as a function of parallax factor and noise standard deviation.

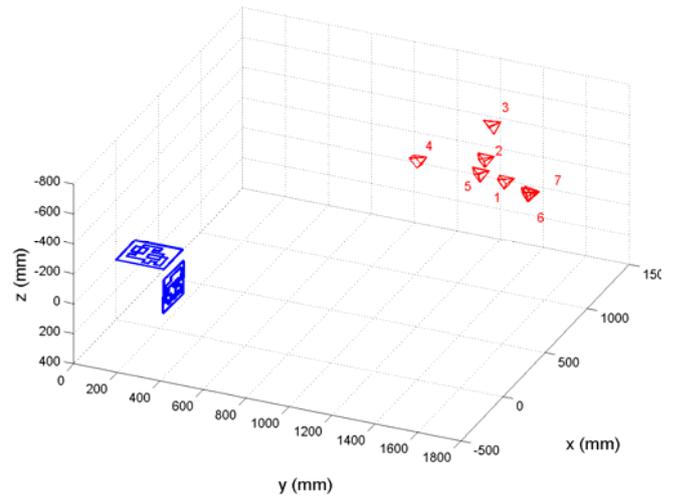

Fig 8(b). Ground-truth camera poses and 3D structure in real tests.

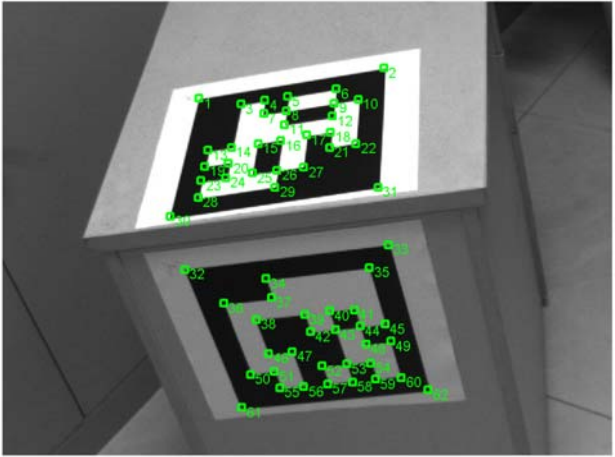

Fig 8(a). Test table and barcode papers with feature points.

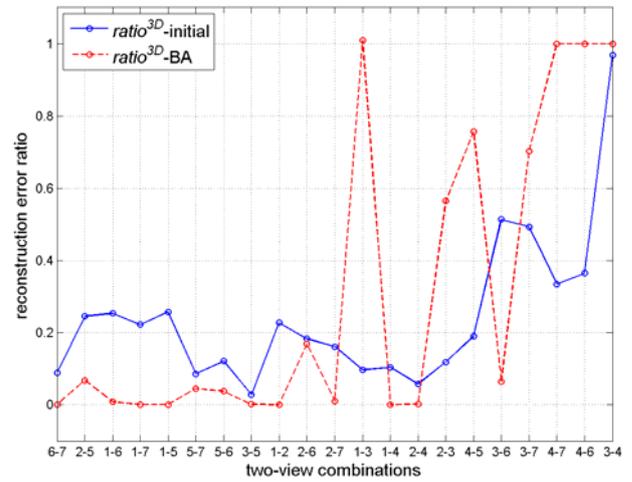

Fig 9. Reconstruction error ratios for 21 two-view combinations in real tests.

factors, and along with further increased parallax factor, the translation discrepancy difference turns to decrease and stops just above zero (see $d$=30m for example).

### 5.2 Analytical Reconstruction Result

According to Proposition 3, the relative depth has been expressed in terms of the pose. It actually gives a new method to analytically reconstruct 3D points. There are two depths existing in (36), one for the left view and the other for the right view. The difference between the two depths is negligible and they are averaged as the depth estimate of the new method. We compare the new method with the traditional 3D point reconstruction method by the DLT-based linear triangulation [13]. The depth of both methods is computed using the identified solution in Section 5.1.

The reconstruction error ratio is defined as

$$ratio^{3D} = \frac{\varepsilon_{anal}^{3D}}{\varepsilon_{trad}^{3D}} \tag{83}$$

where $\varepsilon_{trad}^{3D}$ and $\varepsilon_{anal}^{3D}$ are respectively the 3D point average reconstruction error (RMSE across Monte Carlo runs) of the analytical method and the traditional method. Figure 7 presents the reconstruction error ratio as a function of noise standard deviation and parallax scale. For normal cases with large parallax, the two reconstruction methods perform almost identically, but for small parallax and large noise standard deviation, $ratio^{3D}$ tends to be much less than 1, which means that the analytical method is much better in reconstruction accuracy.

Real tests are carried out to confirm the above reconstruction advantage. Two 2D barcode papers are placed onto mutually-perpendicular faces of a table. Pictures are taken with seven selected poses so that the two barcode pictures are well observed (see Fig. 8). Figure 8(a) gives an example picture of the table and barcode papers, in which 62 feature points are observed, and Fig. 8(b) depicts the seven camera poses relative to the barcode papers.

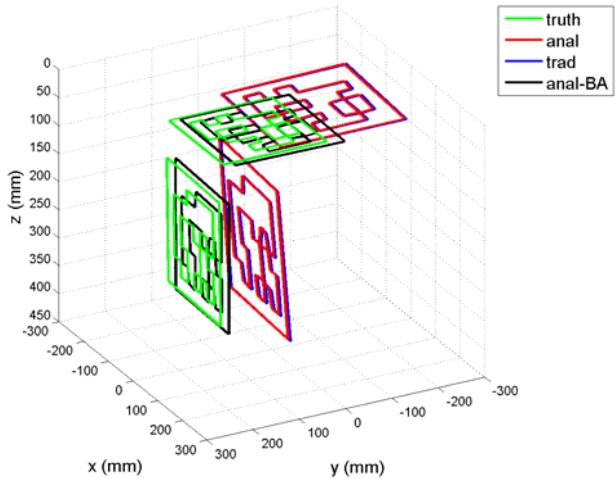

Fig 10(a). Reconstruction results for 3-4 pair of views with largest parallax. Ground truth (in green), analytical reconstruction (in red), traditional reconstruction (in blue) and analytically-initialized BA (in black).

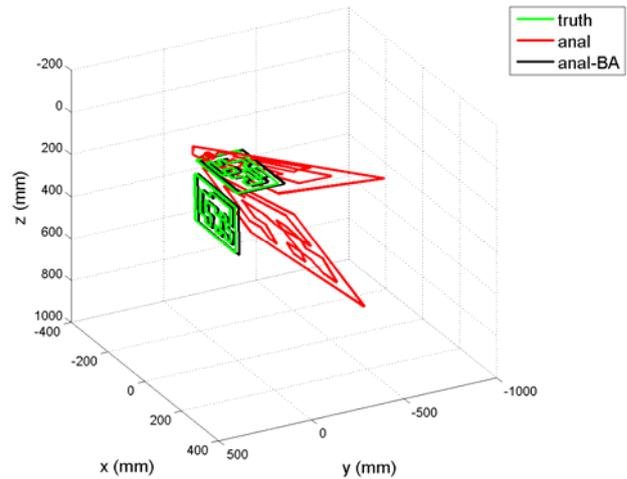

Fig 10(c). Reconstruction results for 1-4 pair of views. Ground truth (in green), analytical reconstruction (in red) and analytically-initialized BA (in black).

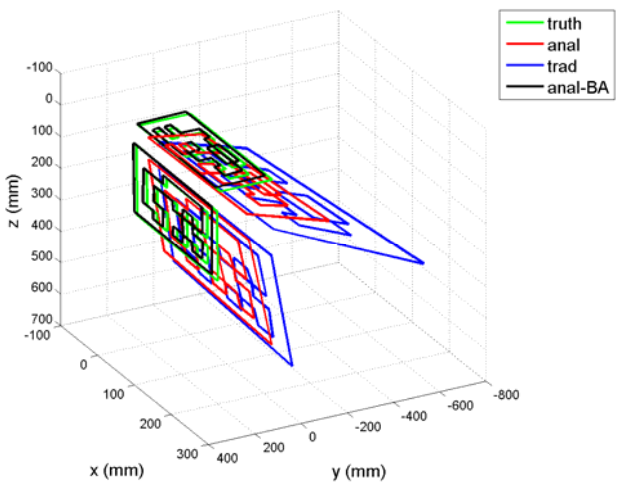

Fig 10(b). Reconstruction results for 4-6 pair of views with second largest parallax. Ground truth (in green), analytical reconstruction (in red), traditional reconstruction (in blue) and analytically-initialized BA (in black).

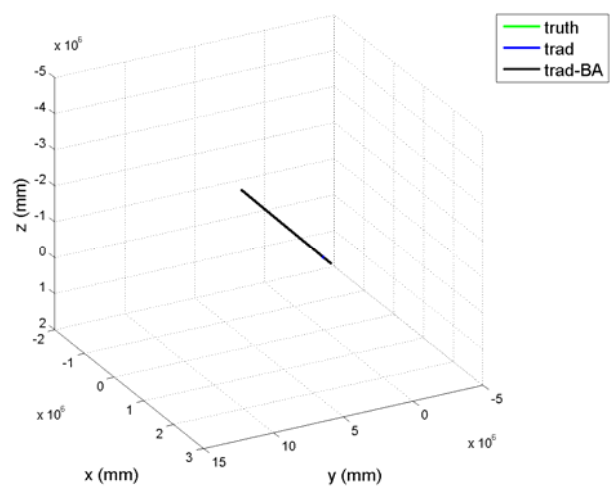

Fig 10(d). Reconstruction results for 1-4 pair of views. Ground truth (in green), traditional reconstruction (in blue) and traditionally-initialized BA (in black).

We use the PnP method [22, 23] to compute the true camera poses and the true 3D points. Specifically, a separate spatial frame is attached to each barcode paper and the PnP method is used to compute the camera poses relative to each barcode paper frame. The ground-truth camera poses and 3D barcode feature point coordinates are obtained by unifying the two spatial frames. The reconstruction error ratios for all $C_7^2 = 21$ combinations (blue circles) are plotted in Fig. 9. Except the 3-4 pair of views, the analytical reconstruction method is overwhelmingly better in reconstruction accuracy.

Considering that the initial pose solution may likely be too coarse to suitable for reconstruction, the bundle adjustment (BA) [13] is used to optimize 3D points and the pose. The BA is respectively initialized by the analytical reconstruction result and the traditional reconstruction result. The reconstruction error ratios after the BA are also presented in red circles in Fig. 9. It shows that the BA does not significantly change the reconstruction error ratio, in other words, the analytical reconstruction method is still much preferred.

Two demonstrating reconstructed barcode papers together with the truth are given in Fig. 10, one is the 3-4 pair of views with the largest parallax, the other one 4-6 pair of views with the second largest parallax. We can see that the two methods perform similarly for the largest parallax case in Fig. 10(a), but the analytical reconstruction method is much better for the second largest parallax case in Fig. 10(b). For 3-4 and 4-6 view pairs, the analytically-initialized BA and traditionally-initialized BA are almost identically good (see Fig. 9), so only the analytically-initialized BA is plotted in Fig. 10. In addition, Figs. 10(c)-(d) give the reconstructed barcode papers for the 1-4 pair of views. The analytical reconstruction initialization leads to a well-behaved BA in Fig. 10(c), while the traditional reconstruction initialization leads to a diverging BA in Fig. 10(d). This can be confirmed by the reconstruction error ratio in Fig. 9, almost zero for the 1-4 view pair.

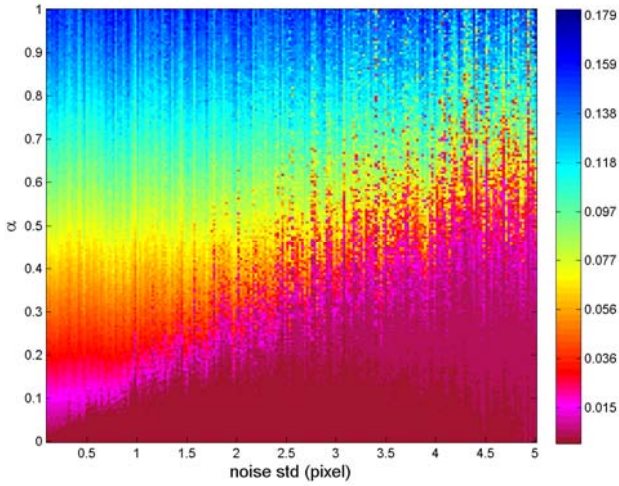

Fig 11(a). PRI as a function of parallax factor and noise standard deviation.

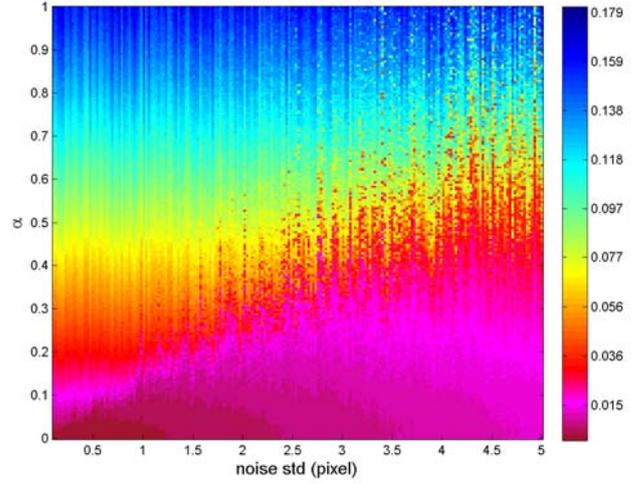

Fig 11(b). $M_3(R)$ as a function of parallax factor and noise standard deviation.

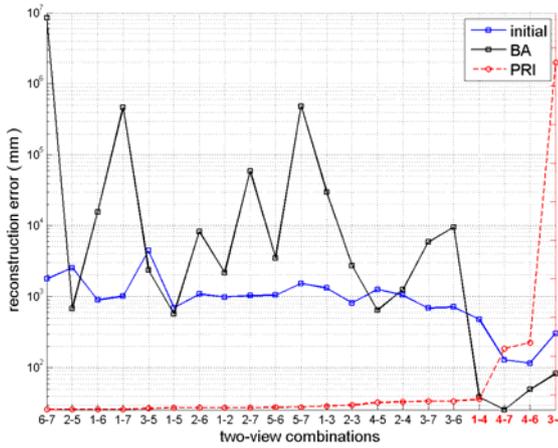

Fig 12(a). Reconstruction errors by analytical method and analytically-initialized BA for 21 two-view combinations in ascending order of PRI.

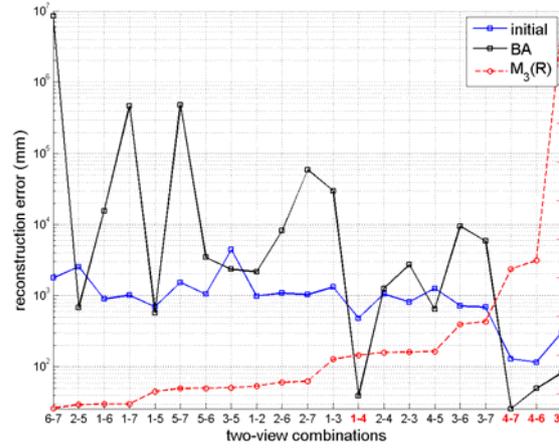

Fig 12(b). Reconstruction errors by analytical method and analytically-initialized BA for 21 two-view combinations in ascending order of $M_3(R)$.

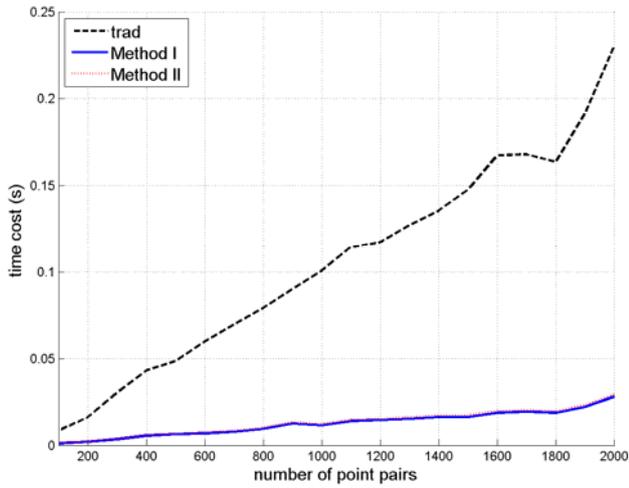

Fig 13(a). Time costs for traditional method, Method I and II, as a function of the number of feature points.

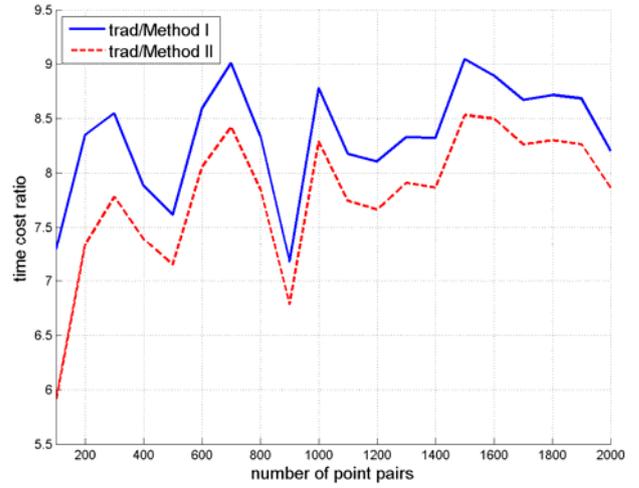

Fig 13(b). Ratios of time cost of traditional method, over Method I and II, as a function of the number of feature points.

## 5.3 Pure Rotation Identification Result

Propositions 4, 5 and 8 actually present two alternative methods to identify pure rotation or infinite 3D points. In Section 4.4, we have compared them for identifying the pure rotation and recommended the method derived from Proposition 8.

Define a pure rotation indicator (PRI) as the average of $M_2(\hat{t})$ in (76) across all feature points

$$PRI(\hat{t}) = avg(|M_2(\hat{t})|) \qquad (84)$$

and identify the motion as a pure rotation if

$$PRI(\hat{t}) < \delta_{pri} \qquad (85)$$

where $\delta_{pri}$ is a threshold. Actually, it is ready to check $|m_2(\hat{t}_1)| = |m_2(\hat{t}_2)|$, so we simply denote it as PRI.

Referring to Fig. 4(b), for the case of pure rotation ($\alpha = 0$) and zero noise, the translation discrepancy is roughly 90 degree. For a fixed nonzero parallax factor, e.g., $\alpha = 0.2$, as the noise standard deviation increases the translation discrepancy also gradually approaches 90 degree. The cases of about 90 degree translation discrepancy roughly forms a triangle at the lower-right corner. Translation estimate with so large discrepancy is usually of little use, so it is preferable to identify all of them by a chosen threshold $\delta_{pri}$.

Figure 11 plots the PRI and $M_3(R)$ with the same simulation settings as Fig. 4. Referring to Fig. 11(a), for the case of pure translation and zero noise, PRI is approximately zero as predicted by Proposition 8. For a fixed nonzero parallax factor, e.g., $\alpha = 0.2$, increasing noise standard deviation tends to reduce the PRI to zero. Notably, the color pattern of PRI much resembles that of the translation discrepancy in Fig. 4(b). This enables us to easily find a threshold to identify the cases with large translation discrepancy. For instance, $\delta_{pri} = 0.015$ corresponding to the magenta color is a good choice to identify the cases with 90 degree translation discrepancy. In contrast, the $M_3(R)$ in Fig. 11(b) has no such a nice property, namely, we cannot find a proper threshold to do the identification. We have tested 3D points with different depths ($d$=30m, 50m and 80m) and found that $\delta_{pri} = 0.015$ is still a good threshold for all of these scenarios.

Figures 12(a)-(b) plot the reconstruction errors by the analytical method and the analytically-initialized BA for 21 two-view combinations, respectively in the ascending order of PRI and $M_3(R)$. The PRI has a much sharper rising than $M_3(R)$. Larger PRI predicts those view pairs (1-4, 4-6, 4-7, 3-4) that have smaller initial reconstruction errors and converged BA results. In comparison, $M_3(R)$ cannot effectively pick the 1-4 view pair out using a threshold, because those view pairs (2-4, 2-3, 4-5, 3-6, 3-7) to the right have diverged BA results.

## 5.4 Computation Cost

Time costs as a function of the number of feature points are shown in Fig. 13. The elapsed time are averaged across 20 Monte Carlo runs. 'Method I' means the two inequalities are employed to identify the right pose solution, and 'Method II' additionally uses the right pose from 'Method I' to analytically reconstruct 3D points. In contrast, the traditional method identifies the right solution by the DLT-based 3D reconstruction and subsequent positive depth check for each of four solutions [13].

Note that the time cost of computing the feature pairs is not taken into account. The matrix $(X_1 \otimes X'_1, \ldots, X_m \otimes X'_m)^T$ in $M_1(R)$ uses the same one as in solving the matrix Q in (42). Figure 13 clearly shows that Methods I and II have almost the same time cost as the analytical reconstruction is very cost-efficient. As compared with the traditional method, they reduce the time cost by about 7-9 times.

## 6 CONCLUSIONS

The well-known essential matrix equation only represents the coplanar relationship in the two-view imaging geometry and loses the important connection among the translation vector and two projection rays. The paper comes up with the PPO (same-side and intersection) constraints for the two-view geometry problem that are proven to be equivalent to the two-view imaging geometry. The complete pose solutions to the essential matrix equation are explicitly derived from the perspective of equation solving. It is shown that the orientation can still be recovered in the pure rotation case. Two inequalities are formulated by simplifying the two new constraints, so as to help directly identify the right solution. It does not need 3D points reconstruction and depth check that are required in traditional methods. The intersection inequality constraint lends itself to a criterion for identifying the pure rotation motion of the camera.

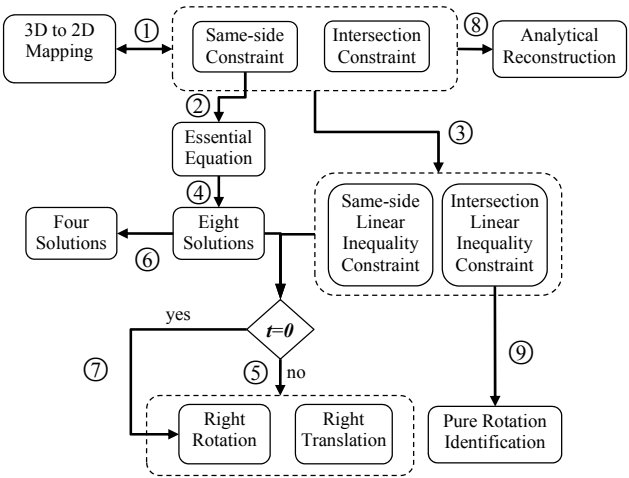

Fig 14. Connections among investigated problems.
①Sufficient and necessary condition; ② Sufficient condition; ③ Linear simplification; ④ Relative poses by SVD; ⑤ Algorithm in Table II; ⑥ Requiring $det(U_Q V_Q) = 1$ or $det(U_Q V_Q) = -1$; ⑦ Obtaining right $R$ only. ⑧ Proposition 3. ⑨ Identification by PRI.

The relationship among the problems discussed in this paper is summarized in Fig. 14 for easy reference. Test results demonstrate the usefulness of the PPO constraints in robustly identifying the right pose solution and pure rotation, as well as in 3D point reconstruction.

## ACKNOWLEDGMENT

Thanks to anonymous reviewers for their constructive comments and Dr. Danping Zou for group talks. The work is funded by National Natural Science Foundation of China (61422311, 61673263, 61503403) and Hunan Provincial Natural Science Foundation of China (2015JJ1021).

## APPENDIX

**Cross Product**: The cross product of two vectors a and b is defined only in three-dimensional space and is denoted by $\mathbf{a} \times \mathbf{b}$. The vector triple product is the cross product of a vector with the result of another cross product, and is related to the dot product by the following formula

$$(\mathbf{a} \times \mathbf{b}) \times \mathbf{c} = (\mathbf{a}^T \mathbf{c}) \mathbf{b} - (\mathbf{b}^T \mathbf{c}) \mathbf{a} \tag{86}$$

and

$$\mathbf{a} \times (\mathbf{b} \times \mathbf{c}) = (\mathbf{a}^T \mathbf{c}) \mathbf{b} - (\mathbf{a}^T \mathbf{b}) \mathbf{c} \tag{87}$$

**Kronecker Product**: If A is an $m \times n$ matrix and B is a $p \times q$ matrix, then the Kronecker product $A \otimes B$ is the $mp \times nq$ block matrix:

$$A \otimes B = \begin{bmatrix} a_{11} B & \cdots & a_{1n} B \\ \vdots & \ddots & \vdots \\ a_{m1} B & \cdots & a_{mn} B \end{bmatrix} \tag{88}$$

**Vectorization**: the vectorization of an $m \times n$ matrix A, denoted $vec(A)$, is the $mn \times 1$ column vector obtained by stacking the columns of the matrix A on top of one another:

$$vec(A) = [a_{11}, \ldots, a_{m1}, a_{12}, \ldots, a_{m2}, \ldots, a_{1n}, \ldots, a_{mn}]^T \tag{89}$$

Here, $a_{ij}$ represents the element of row-$i$ and column-$j$. The vectorization is frequently used together with the Kronecker product to express matrix multiplication as a linear transformation on matrices. In particular,

$$vec(ABC) = (C^T \otimes A) vec(B) \tag{90}$$

for matrices A, B, and C of dimensions $k \times l$, $l \times m$, and $m \times n$.